\documentclass{article}

\usepackage[preprint]{styles}
\usepackage[numbers]{natbib}
\usepackage{graphicx} % Required for inserting images
\usepackage{authblk}
\usepackage[utf8]{inputenc} % allow utf-8 input
\usepackage[T1]{fontenc}    % use 8-bit T1 fonts
\usepackage{hyperref}       % hyperlinks
\usepackage{url}            % simple URL typesetting
\usepackage{booktabs}       % professional-quality tables
\usepackage{amsfonts}       % blackboard math symbols
\usepackage{nicefrac}       % compact symbols for 1/2, etc.
\usepackage{microtype}      % microtypography
\usepackage{xcolor}         % colors
\usepackage{amsmath}
\usepackage{amssymb}
\usepackage{mathtools}
\usepackage{amsthm}
\usepackage{mdframed}
\usepackage{multirow}
\usepackage{makecell}  
\usepackage{titlesec}
\usepackage{algorithm}  %
\usepackage{algorithmic}%
\usepackage[capitalize,noabbrev]{cleveref}
\usepackage{microtype}
\usepackage{wrapfig}
\usepackage{graphicx}
\usepackage{subcaption}
\usepackage{booktabs} % for professional tables
\usepackage{amsfonts}
\usepackage{amssymb}

\usepackage{enumitem}
\renewcommand{\citet}[1]{\cite{#1}}
\renewcommand{\cite}[1]{\citep{#1}}

\title{Conceptual Framework Toward Embodied Collective Adaptive Intelligence}

\author[1]{Fan Wang}
\author[1]{Shaoshan Liu}
\affil[1]{Shenzhen Institute of Artificial Intelligence and Robotics for Society, China}

\date{May 2025}

\begin{document}

\maketitle

\begin{abstract}
Collective Adaptive Intelligence (CAI) represent a transformative approach in embodied AI, wherein numerous autonomous agents collaborate, adapt, and self-organize to navigate complex, dynamic environments. By enabling systems to reconfigure themselves in response to unforeseen challenges, CAI facilitate robust performance in real-world scenarios. This article introduces a conceptual framework for designing and analyzing CAI. It delineates key attributes including task generalization, resilience, scalability, and self-assembly, aiming to bridge theoretical foundations with practical methodologies for engineering adaptive, emergent intelligence. By providing a structured foundation for understanding and implementing CAI, this work seeks to guide researchers and practitioners in developing more resilient, scalable, and adaptable AI systems across various domains.
\end{abstract}

\section{Prospectives and Challenges}

This section explores the conceptual and practical foundations required for building embodied collective adaptive intelligence (CAI). We examine how principles of emergence, modular simplicity, and system-level adaptivity intersect, and what computational and architectural challenges must be overcome to translate these ideas into scalable, real-world AI systems.

\textbf{Emergence and Collectivity in Complex Systems}:  
The idea that “more is different” \cite{anderson1972more} has long explained how complex behaviors can arise from the interactions of simple components. This principle—central to phenomena in physics, biology, and artificial intelligence—also underpins recent advances in deep learning. In collective systems such as neural modular architectures and swarm intelligence, researchers have begun to bridge local interaction rules with emergent, system-wide capabilities \cite{ha2022collective}. These examples show how collectivity can serve as both a functional and scalable design paradigm.

\textbf{Simplicity as a Design Principle}:  
Occam’s Razor reminds us that simplicity can drive generalization. In machine learning, simpler models often avoid overfitting and adapt better to new inputs \cite{domingos1999role}. This insight applies to collective systems as well: architectures built from small, reusable, parameter-efficient modules may outperform large monolithic models. By promoting modularity and inductive bias, such systems achieve flexibility without sacrificing robustness—especially important when scaling across diverse tasks and domains.

\textbf{Why Collective Intelligence Stresses Today’s Hardware}:  
Despite their promise, collective systems face a fundamental mismatch with existing computing infrastructure. Unlike feedforward models that scale linearly on GPUs, collective architectures often rely on recursive message passing and interdependent modules, making them difficult to parallelize \cite{ha2022collective}. As a result, these models can be more memory- and compute-intensive—even if they use fewer total parameters. Unlocking their potential will require co-designing new hardware and software stacks optimized for decentralized, interaction-heavy computation.

\textbf{From Collective Intelligence to Collective Adaptivity}:  
Collective adaptivity goes a step further, enabling AI systems to reorganize themselves—changing their structure, behavior, or scale—at inference time. This concept, well-established in complex systems and multi-agent software engineering \cite{fernandez2013description,capodieci2016artificial}, remains underexplored in AI. We argue that embodied AI systems capable of adapting their topology, communication patterns, or component roles on the fly could break free from the constraints of static training-time architectures.

Early signs of this potential appear in domains like large language models—where emergent task coordination can arise without explicit programming \cite{hongmetagpt}—and in autonomous agents navigating open-ended environments \cite{bauer2023human}. Still, true collective adaptivity remains rare. Most current systems lack the robustness to degrade gracefully under partial failure, or the flexibility to reallocate resources dynamically in response to context. Bridging this gap between theoretical adaptivity and practical deployment is a major unsolved challenge.

In this article, we formally define \textit{embodied collective adaptive intelligence (CAI)}, and outline its core attributes, such as scalability, decentralization, and self-organization. Our goal is to offer a conceptual framework that supports not just analysis, but the engineering of real-world systems capable of emergent, adaptive behavior.

\section{Collective Adaptive Intelligence: Conceptual Definitions}

Building on the prior discussion of emergence, modularity, and adaptivity, we now formalize the concept of \textit{Collective Adaptive Intelligence (CAI)}. This section offers a framework for distinguishing between different levels of collective intelligence and introduces key features, adaptation, resilience, scalability, and self-assembly—that distinguish CAI from traditional multi-agent systems. These distinctions help ground future system design in both theory and practice.

\begin{figure*}[!htbp]
\centering
\begin{subfigure}[b]{0.49\textwidth}
    \centering 
    \includegraphics[width=0.99\textwidth]{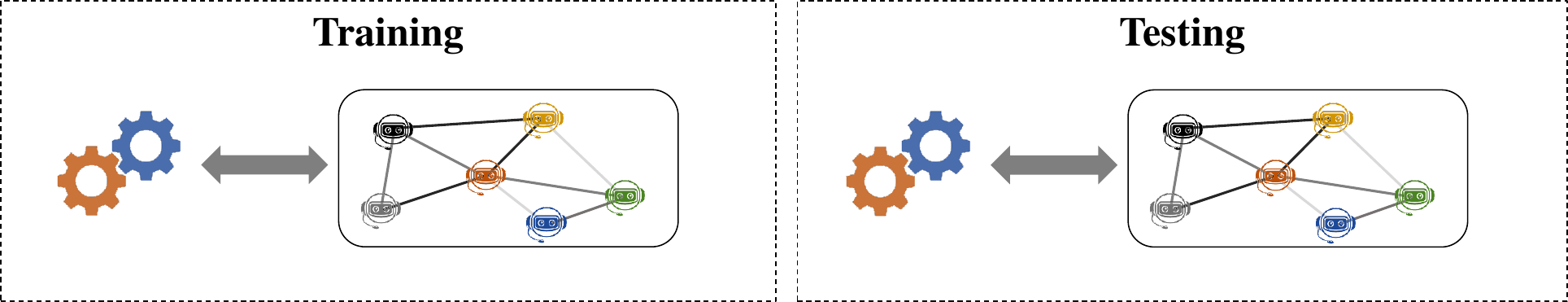}
    \caption{Specialist Collective Intelligence}
    \label{fig:CAS_1}
\end{subfigure}
\begin{subfigure}[b]{0.49\textwidth}
    \centering 
    \includegraphics[width=0.99\textwidth]{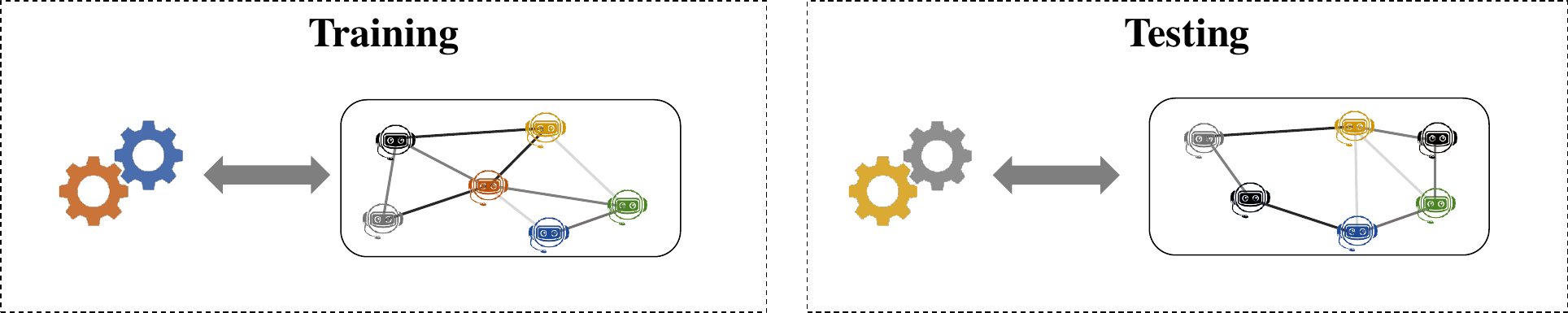}
    \caption{Generalist Collective Intelligence}
    \label{fig:CAS_2}
\end{subfigure}
\\
\begin{subfigure}[b]{0.49\textwidth}
    \centering 
    \includegraphics[width=0.99\textwidth]{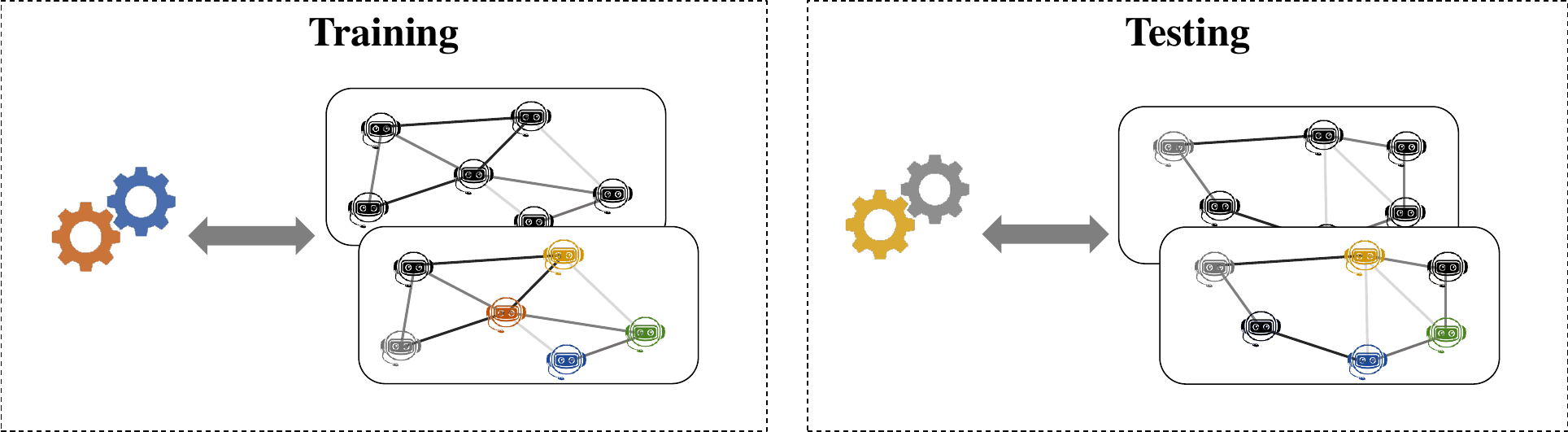}
    \caption{Task and topology adaption}
    \label{fig:CAS_3}
\end{subfigure}
\begin{subfigure}[b]{0.49\textwidth}
    \centering 
    \includegraphics[width=0.99\textwidth]{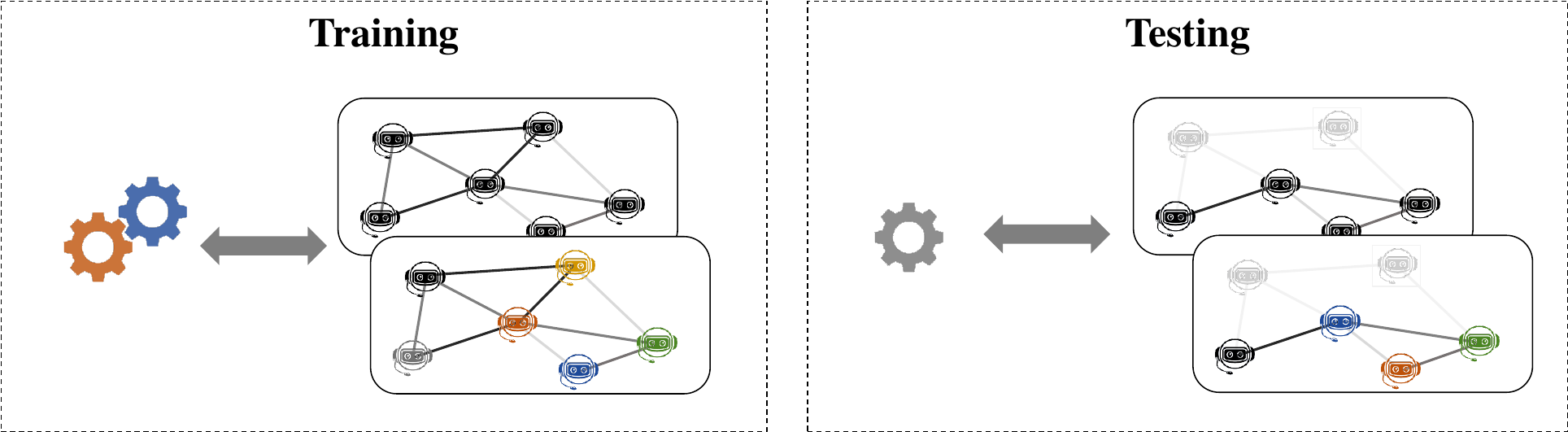}
    \caption{Collective Resilience}
    \label{fig:CAS_4}
\end{subfigure}
\\
\begin{subfigure}[b]{0.49\textwidth}
    \centering 
    \includegraphics[width=0.99\textwidth]{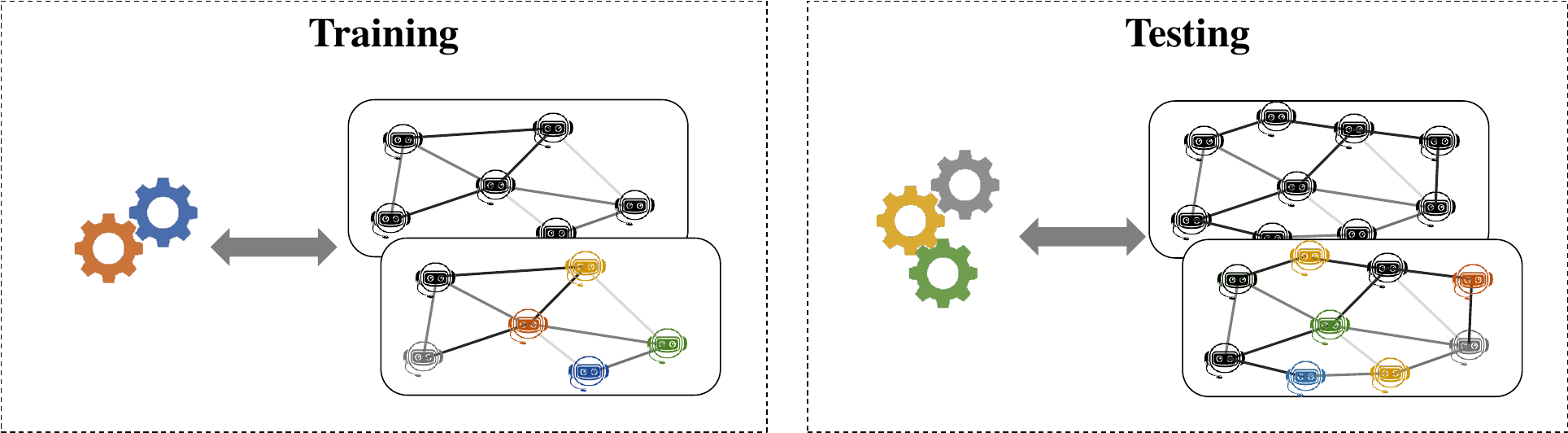}
    \caption{Collective Scalability}
    \label{fig:CAS_5}
\end{subfigure}
\begin{subfigure}[b]{0.49\textwidth}
    \centering 
    \includegraphics[width=0.99\textwidth]{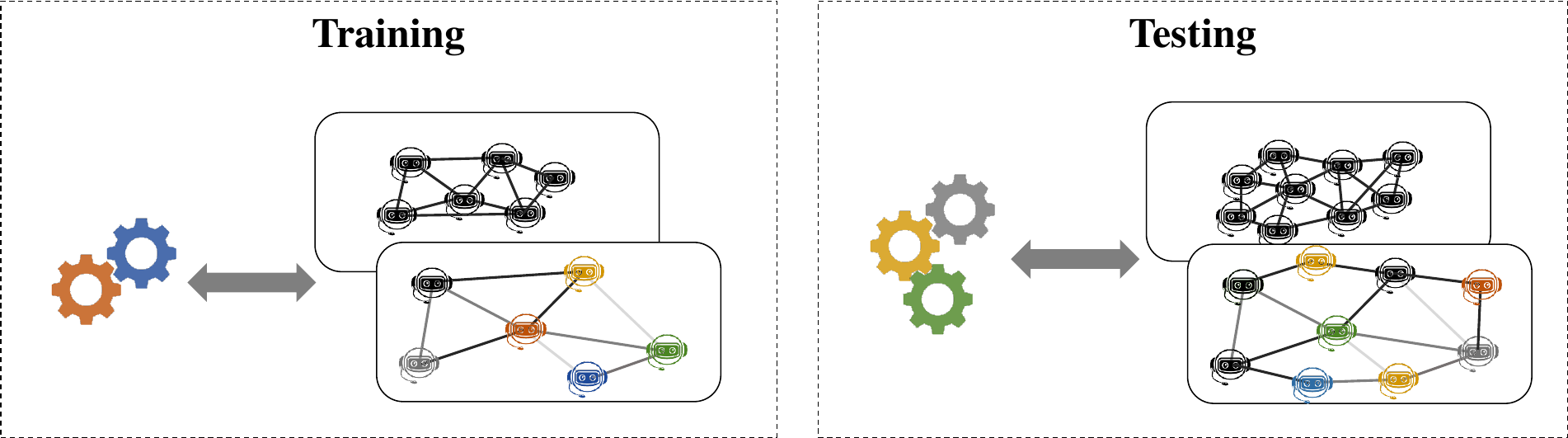}
    \caption{Self-Assembly}
    \label{fig:CAS_6}
\end{subfigure}
\caption{Collective intelligences and different features of collective adaptive intelligences.
Specialist: the swarm is limited to handling only one specific task and maintaining a fixed topology;
Generalist: the swarm possesses the ability to instantaneously tackle various tasks and adapt to different predefined topologies;
Task and topology generalization:  the swarm adapts to diverse tasks and collaborative patterns; Collective resilience: the swarm reconfigures its task allocation to compensate for agent failures or loss of functionality.; Collective scalability: larger swarms enable the solution of increasingly complex tasks during testing; Self-assembly: the swarm autonomously reorganizes its collaborative structure to adapt to new tasks.}
\label{fig:CAS}
\end{figure*}

We first consider a conceptual definition of collective adaptive intelligent (CAI) system as shown in Figure~\ref{fig:CAS}. Canonical Multi-Agent Systems (MAS) are trained on single tasks and deployed solely on the training task. In traditional settings, the coordination patterns among components are predefined, which is regarded as specialist collective intelligence (SCI). However, if the MAS is pretrained on a diverse set of numerous tasks, it can potentially be deployed not only on a broad range of in-distribution tasks but also on some out-of-distribution (OOD)tasks. In such cases,it is considered as a generalist collective intelligence (GCI). CAI should possess at least the following features (as shown in \cref{fig:CAS}):
\begin{itemize}
\item \emph{Task and topology adaption}: Achieving generalization across diverse out-of-distribution (OOD) tasks and varying topologies, including different coordination patterns.
\item \emph{Collective Resilience}: Demonstrating the ability to re-adapt and complete OOD tasks even when some components are removed from the collective.
\item \emph{Collective Scalability}: Showing the capability to re-adapt and handle more complex OOD tasks when new components are added to the collective.
\item \emph{Self-Assembly}: The ability to self-organize from any initial topology into an appropriate structure to accomplish OOD tasks.
\end{itemize}

\textbf{Adaptation (CAI) vs. Zero-Shot Generalization (GCI)}: The difference between adaptation and zero-shot generalization is often blurred in discussions of model generalization. Yet, this distinction becomes critical in settings governed by partially observable Markov decision processes (POMDPs), where zero-shot generalization is fundamentally limited, as shown in prior work on meta-learning and general-purpose agents \cite{thrun1998learning,fan2025putting}. In collective systems, where each agent only perceives a fragment of the environment and must rely on limited inter-agent communication—generalization is rarely achievable without some degree of adaptation. The need to integrate partial and dynamic information at inference time makes adaptation a more viable and often necessary strategy. Crucially, attempts to maximize zero-shot generalization can work against the flexibility required for robust performance in such decentralized, communication-bound systems. Therefore, designing truly scalable collective intelligence systems demands prioritizing adaptive mechanisms over fixed generalization.

\section{A Conceptual Framework for Collective Adaptive Agents}

Building on the foundational properties of collective adaptivity outlined in the previous section, we now introduce a conceptual framework for constructing \textit{Collective Adaptive Agents (CAAs)}—the foundational units of Collective Adaptive Intelligence (CAI).

\begin{figure}[!htbp]
\begin{subfigure}[b]{0.49\textwidth}
    \centering 
    \includegraphics[width=0.99\textwidth]{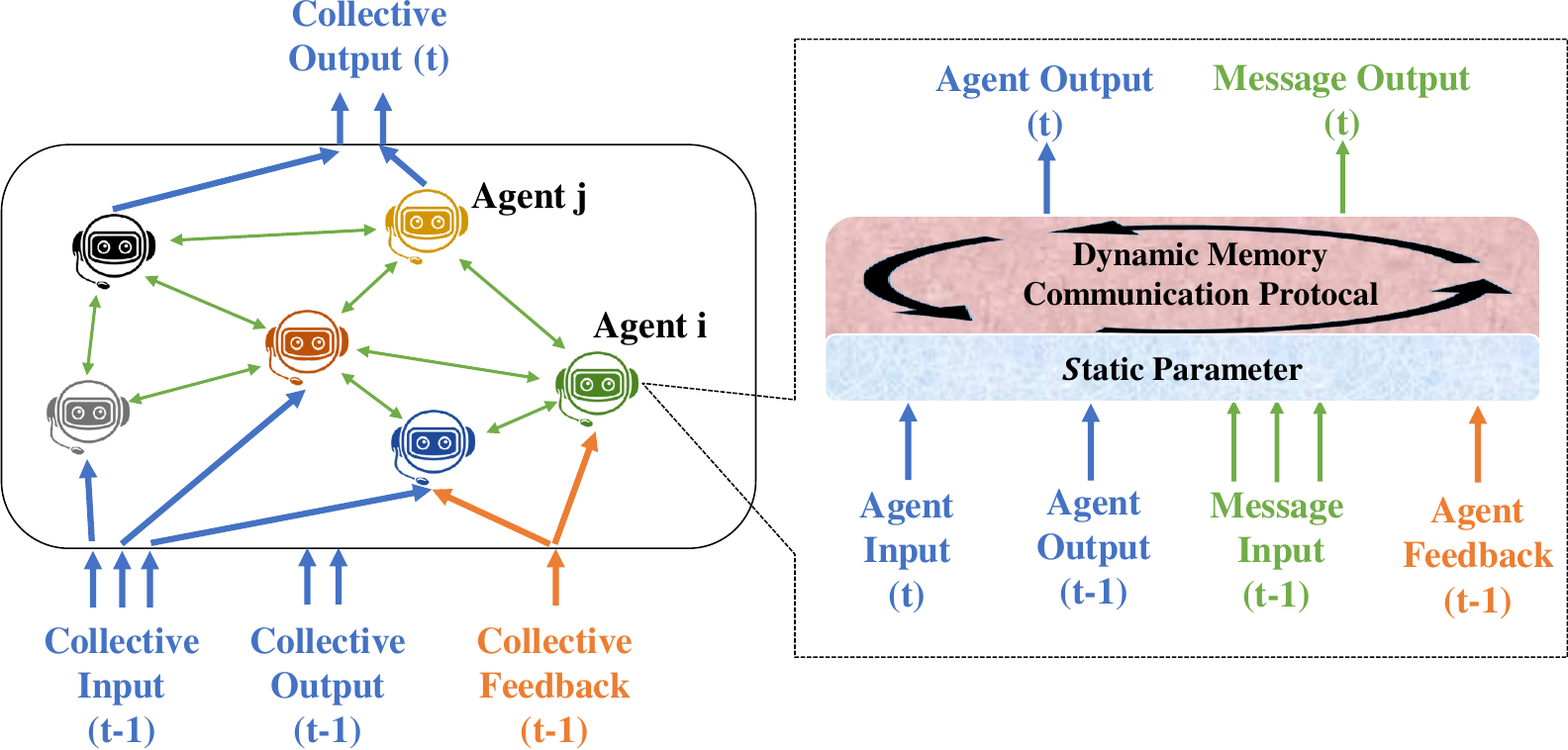}
    \caption{A CAA formulation}
    \label{fig:CAS_7}
\end{subfigure}
\begin{subfigure}[b]{0.49\textwidth}
    \centering 
    \includegraphics[width=0.99\textwidth]{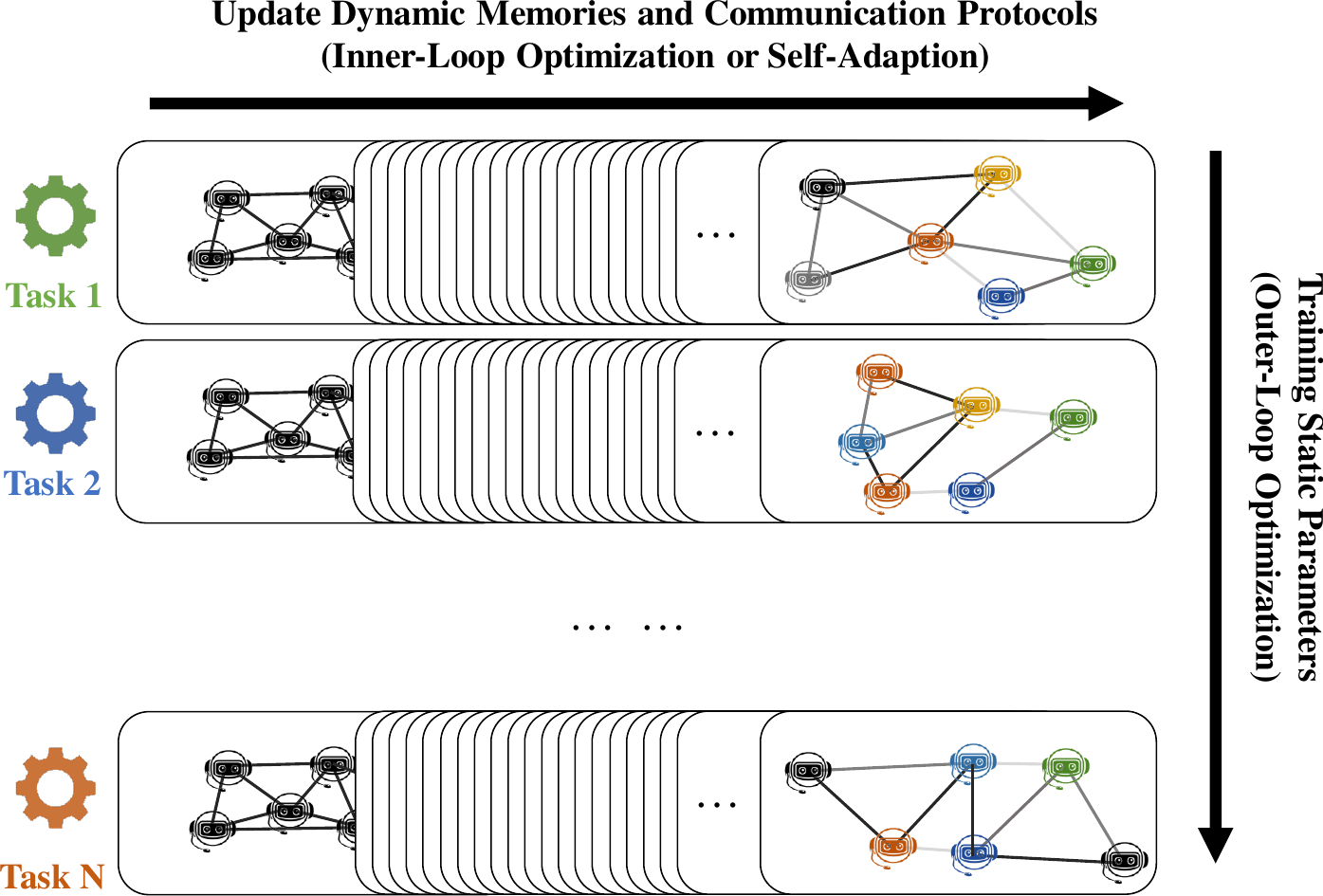}
    \caption{Training Process}
    \label{fig:CAS_8}
\end{subfigure}
\caption{An illustration of the formulation of Collective Adaptive Agent (CAA) and its training process.}
\label{fig:Training}
\end{figure}

A well-structured conceptual framework for collective adaptive agents (CAAs) must incorporate the following core attributes to ensure robustness and scalability in distributed systems:  
\begin{itemize}
    \item \emph{Decentralization}:  Agents operate independently within a decentralized system, eliminating reliance on centralized control mechanisms. 
    \item \emph{Self-Adaptation}: Agents dynamically adjust their input-output mappings and messaging strategies based on historical data. This capability is realized through real-time adaptation of internal memory structures or learning algorithms.
    \item \emph{Self-Assembly and Collective Scalability}: Agents autonomously modify their communication protocols to selectively accept messages from specific peers and route outputs to relevant recipients. 
\end{itemize}

As depicted in Figure~\ref{fig:Training}, to fulfill the aforementioned requirements, a collective adaptive agent (CAA) must account for historical interaction data and feedback signals to dynamically adapt its role within a collective system, thereby enabling effective task generalization across diverse scenarios. The proposed CAA architecture incorporates three distinct parameter sets: 
\begin{itemize}
    \item \textbf{Static Parameters}: These remain fixed during inference to maintain structural consistency. 
    \item \textbf{Dynamic Memories}: These are updated in real-time to enable self-adaptation to novel contexts or roles.
    \item \textbf{Communication Protocols}: These alters coordination strategies among agents to align with evolving task demands.
\end{itemize}
This complex system can be optimized by drawing inspiration from meta-learning principles~\cite{thrun1998learning}, which results in a hierarchical learning process (Figure~\ref{fig:CAS_8}):
\begin{itemize}
    \item \textbf{Outer-Loop Optimization}: Searches for optimal static parameters to define the agent’s foundational capabilities and adaption functionalities.  
    \item \textbf{Inner-Loop Adaptation}: Dynamically updates dynamic memory and communication protocols during inference to ensure rapid role-switching and task-specific coordination.  
\end{itemize}

Together, these design principles highlight a key insight: scalable collective intelligence emerges not from static specialization, but from dynamic, decentralized agents capable of continual adaptation in structure, behavior, and communication.

\section{Limitations and Future Challenges}

While collective intelligence offers a compelling paradigm for scalable embodied AI, current computational architectures are poorly suited to its decentralized, communication-heavy nature. A key challenge lies in developing Collective Adaptive Agents (CAAs) that can be trained efficiently at small scales yet generalize through self-organization at deployment. Early signs from LLMs and open-ended agents are promising, but realizing scalable, adaptive collectivity demands co-evolution of algorithms, architectures, and infrastructure. \textbf{The path forward lies not in scaling monolithic models, but in designing systems where intelligence emerges through adaptive interaction.}

\bibliographystyle{plainnat}
\bibliography{main}

\end{document}